\documentclass{article}
\usepackage{textcomp}

\usepackage{PRIMEarxiv}

\usepackage[utf8]{inputenc} %
\usepackage[T1]{fontenc}    %
\usepackage{hyperref}       %
\usepackage{url}            %
\usepackage{booktabs}       %
\usepackage{amsfonts}       %
\usepackage{nicefrac}       %
\usepackage{microtype}      %
\usepackage{lipsum}
\usepackage{tabularx}
\usepackage{fancyhdr}       %
\usepackage{graphicx}       %
\usepackage{natbib}
\usepackage[most]{tcolorbox}
\usepackage{multirow}
\usepackage{threeparttable}
\graphicspath{{media/}}     %
\usepackage{algorithm}
\usepackage{algorithmic}
\usepackage{amsmath}
\usepackage{wrapfig}
\usepackage{booktabs}
\usepackage{pifont}

\title{PDE-Agent: A toolchain-augmented multi-agent framework for PDE solving}

\NewDocumentCommand{\hongru}
{ mO{} }{\textcolor{red}{\textsuperscript{\textit{Hongru}}\textsf{\textbf{\small[#1]}}}}

\NewDocumentCommand{\qixuan}
{ mO{} }{\textcolor{green}{\textsuperscript{\textit{qixuan}}\textsf{\textbf{\small[#1]}}}}

\NewDocumentCommand{\yimin}
{ mO{} }{\textcolor{blue}{\textsuperscript{\textit{yimin}}\textsf{\textbf{\small[#1]}}}}

\begin{document}

\maketitle
\vspace{-7em}
\begin{center}
Jianming Liu$^{*1,2}$, Ren Zhu$^{*1,3}$, Jian Xu$^{1,3}$, Kun Ding$^{1}$, Xu-Yao Zhang$^{1,3}$, Gaofeng Meng$^{1,3}$, Cheng-Lin Liu$^{1,3}$\textsuperscript{\ding{41}} 
\end{center}

\begin{center}
\textsuperscript{1}MAIS, Institute of Automation, Chinese Academy of Sciences\\
\textsuperscript{2}School of Advanced Interdisciplinary Sciences, University of Chinese Academy of Sciences \\
\textsuperscript{3}School of Artificial Intelligence, University of Chinese Academy of Sciences \\
\end{center}

\vspace{-1em}

\begingroup
 \let\thefootnote\relax
\footnotetext{$^*$ These authors contributed equally to this work. \textsuperscript{\ding{41}}Corresponding Author.}   
\endgroup

\vspace{3em}

\begin{abstract}
Solving Partial Differential Equations (PDEs) is a cornerstone of engineering and scientific research. Traditional methods for PDE solving are cumbersome, relying on manual setup and domain expertise. While Physics-Informed Neural Network (PINNs) introduced end-to-end neural network-based solutions, and frameworks like DeepXDE further enhanced automation, these approaches still depend on expert knowledge and lack full autonomy. In this work, we frame PDE solving as tool invocation via LLM-driven agents and introduce PDE-Agent, the first toolchain-augmented multi-agent collaboration framework, inheriting the reasoning capacity of LLMs and the controllability of external tools and enabling automated PDE solving from natural language descriptions. PDE-Agent leverages the strengths of multi-agent and multi-tool collaboration through two key innovations: (1) A Prog-Act framework with graph memory for multi-agent collaboration, which enables effective dynamic planning and error correction via dual-loop mechanisms (localized fixes and global revisions). (2) A Resource-Pool integrated with a tool-parameter separation mechanism for multi-tool collaboration. This centralizes the management of runtime artifacts and resolves inter-tool dependency gaps in existing frameworks. To validate and evaluate this new paradigm for PDE solving , we develop PDE-Bench, a multi-type PDE Benchmark for agent-based tool collaborative solving, and propose multi-level metrics for assessing tool coordination. Evaluations verify that PDE-Agent exhibits superior applicability and performance in complex multi-step, cross-step dependent tasks. This new paradigm of toolchain-augmented multi-agent PDE solving will further advance future developments in automated scientific computing. Our source code and dataset will be made publicly available.
\end{abstract}

\section{Introduction}

Partial Differential Equations (PDEs) are fundamental for modeling intricate phenomena in science and engineering, with applications in fluid dynamics \cite{howard2015data}, heat transfer \cite{zobeiry2021physics}, and quantum mechanics, underpinning our understanding of physical systems. Efficiently solving PDEs remains a critical bottleneck.
Conventional numerical methods, finite element (FEM) \cite{nikishkov2004introduction}, finite difference (FDM) \cite{causon2010introductory}, finite volume (FVM) \cite{eymard2000finite}, rely on mesh-based discretization, incur high computational costs, and face the curse of dimensionality.
Physics-Informed Neural Networks (PINNs) \cite{raissi2019physics, cuomo2022scientific, torres2024survey}, a promising deep learning approach, constrain neural networks to minimize PDE residuals, using automatic differentiation for precise derivative computation to avoid truncation errors and demonstrate potential for overcoming dimensional limitations. Frameworks like DeepXDE \cite{lu2021deepxde} have advanced their practical use.
Yet current methods still require significant domain expertise and programming skills, limiting their broader scientific utility. Thus, developing automated solving frameworks to reduce entry barriers and boost efficiency is critical to advancing the field.

%


\begin{wrapfigure}{R}{0.46\linewidth} 
\centering 
\includegraphics[width=0.95\linewidth]{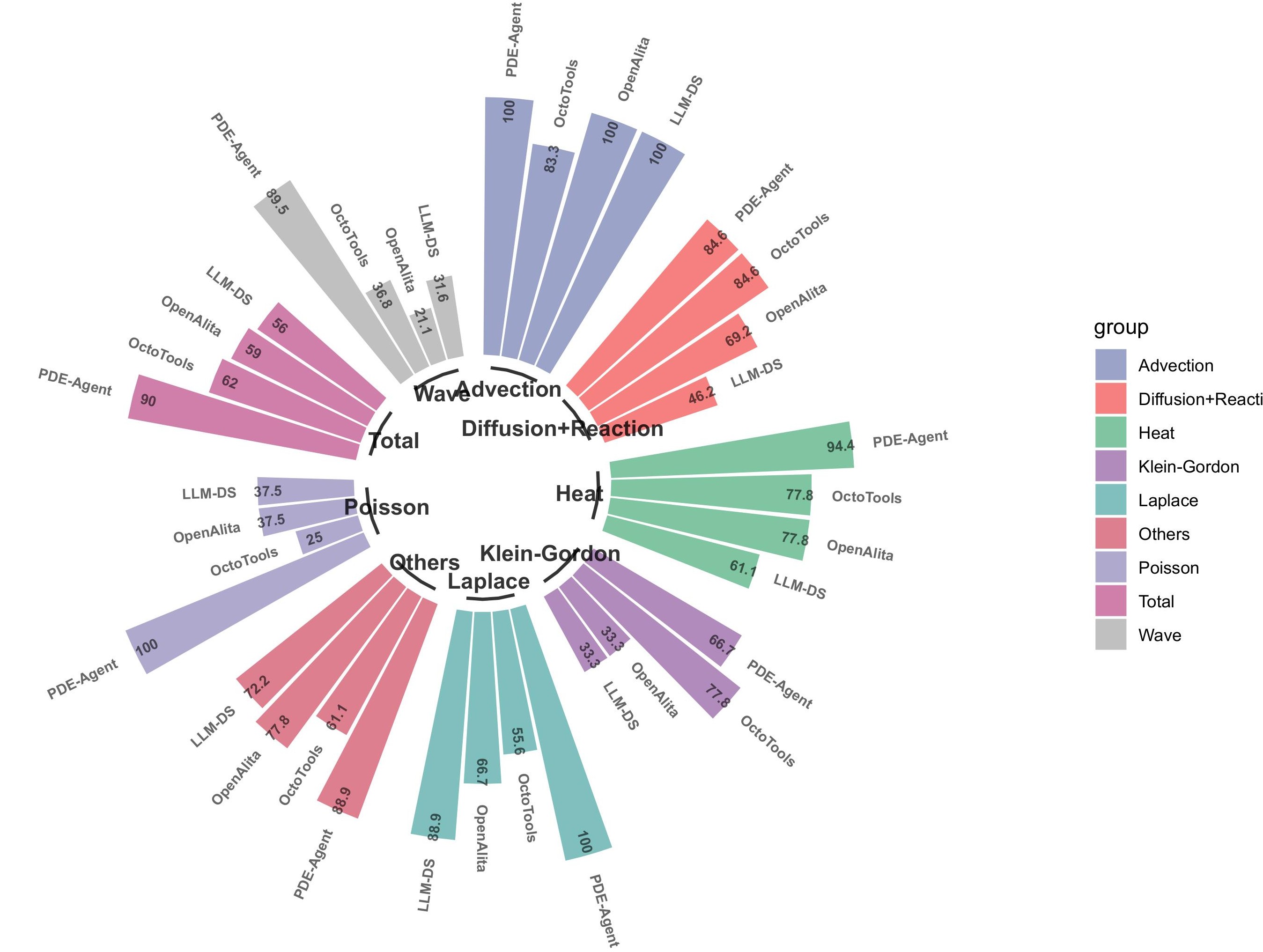}
\caption{The competitive and stable performance across diverse PDE problems.}
\label{fig:intro}
\end{wrapfigure}

Large Language Models (LLMs) \cite{achiam2023gpt, brown2020language} have shown strong reasoning and problem-solving abilities in general domains, with LLM-driven agents increasingly applied to disciplinary problems and automated scientific exploration \cite{ghafarollahi2025sciagents, schmidgall2025agentrxiv, ai4science2023impact}, notably in text understanding \cite{el2021automatic}, code generation \cite{nakano2021webgpt, zhang2024codeagent}, and interdisciplinary QA \cite{kim2024mdagents}. However, their use in specialized scientific computing faces critical challenges: insufficient domain knowledge (limited coverage and poor understanding of specialized symbols/descriptions), struggles with complex computational tasks (declining performance and hallucinations), and reliance on statistical patterns in training data rather than true comprehension of physical essence, leading to unreliable results.

%
Analogous to humans, tool-using is key to expanding cognitive boundaries and boosting capabilities \cite{gou2024critic}. Inspired by this insight, LLM agents collaborating on tool use can bridge their capability and knowledge gaps in complex scientific computing \cite{zhang2023evaluating}. This multi-tool collaborative approach offers a promising new paradigm for automated PDE solving: it automatically converts high-level natural language PDE descriptions into executable task planning, tool scheduling, and response generation, boosting problem-solving automation/efficiency while enhancing solution interpretability and robustness.

Meanwhile, current research on tool-using agent \cite{chen2025optmetaopenfoam, lu2025octotools, qiu2025alita} largely relies on end-to-end metrics (e.g., Pass@K, BLEU \cite{papineni2002bleu}) measuring task success or final output quality. While these capture overall task completion, they overly focus on static outcomes and neglect dynamic processes, failing to quantify fine-grained dimensions of tool-using ability. Notably, in scientific computing, with strong causal dependencies and high fault tolerance needs, this outcome-centric approach cannot diagnose bottlenecks or support optimization/verification of coordination strategies. Thus, there is an urgent need for multi-dimensional process evaluations that transcend end-to-end metrics to scientifically deconstruct agents' tool-using performance.


To this end, we present a multi-type PDE benchmark for tool-collaboration invocation - PDE-Bench, which contains differential equations with a total of approximately 100 test cases, along with a general multi-tiered metrics that can comprehensively assess tool invocation at the global, logical, and local-detial levels.
To demonstrate the generality of our approach, we evaluate it on this benchmark and concurrently present PDE-Agent, an innovative multi-agent framework equipped with modular PDE-toolkits that we have rigorously encapsulated for automated PDEs solving. The architecture integrates task-aware decomposition, tool-oriented execution, and self-refining collaboration strategies. 
%

PDE-Agent establishes an autonomous, closed-loop architecture from natural language problem descriptions to PDE solutions. 
To achieve better multi-agent collaboration, we further propose a Prog-Act, a Progressive reasoning and Acting approach. Prog-Act is a semi-dynamic planning method that enables active assistance-seeking, fixed-point verification and escalating collaboration  through multi-agent design, and realizes error localization and error tracing by tool-chain constructing dynamically.
In addition, we propose a resource-pool design to address the issue of implicit parameter passing in multi-tool collaboration. Fig.\ref{fig:intro} provides a task-level accuracy overview, demonstrating that our approach achieves competitive and stable performance across diverse PDE problems; detailed statistics are deferred to Section \ref{sec:experiments}.

To summarize, the critical contributions of our work are the following: 
\begin{enumerate}
\item We introduce PDE-Agent, the first LLM-driven multi-agent framework equipped with modular PDE-toolkits - rigorously encapsulated for automated PDEs solving. 
\item We develop PDE-Bench, a manually curated multi-type PDE benchmark for tool-collaborative invocation, comprising differential equations across approximately 100 cases. Alongside this dataset, we propose general multi-level evaluation metrics encompassing global, logical, and local-detail-level assessments for comprehensive tool invocation evaluation.
\item We further present Prog-Act, a Progressive reasoning and Acting approach, and a resource-pool design. These innovations enhance multi-agent collaboration and multi-tool coordination, directly improving the robustness and efficiency of automated PDE solving. 
\item Finally, we comprehensively evaluated and compared agent and tool invocation methods, and the experimental results demonstrate the effectiveness of PDE-Agent.
\end{enumerate}

\section{Related Work}
\subsection{Automation and Intelligence in PDE Solving}

As a critical tool for describing physical systems, automated solving of PDEs remains a key challenge in scientific computing. Traditional numerical methods (e.g., FEM, FDM) \cite{zienkiewicz2013finite, leveque2007finite} rely on manual mesh generation \cite{MAVRIPLIS1996417} and discretization tuning, limiting automation for complex geometries or multi-physics problems. Deep learning has shifted focus to intelligent PDE solving, with PINNs \cite{raissi2019physics} emerging as a leading framework \cite{cuomo2022scientific, torres2024survey, zhongkai2024pinnacle}: by embedding PDE constraints into the loss function, PINNs avoid mesh-based discretization, enabling end-to-end PDE solution and marking a key advance in automation.
Neural-network-based tools like NeuralPDE \cite{rackauckas2017differentialequations}, PyDEns \cite{koryagin2019pydens}, and DeepXDE build on this, advancing automation via unified interfaces for defining PDEs, configuring networks, and optimizing training—lowering barriers for non-experts to use PINNs.


However, existing PINNs-based methods still face key limitations for fully automating PDE solving: PDEs and their constraints often require manual coding (conflicting with ``fully automated'' goals), and adjustments to network structures or training strategies remain heavily human-dependent, lacking systematic collaborative mechanisms. To address these, our work inherits PINNs’ core ideas and draws on DeepXDE, encapsulating PDE-toolkits to advance automation. To our knowledge, it is the first framework to automate PDE solution via coordinated multi-agent tool invocation, independent of expert intervention.

\subsection{Tool-Augmented LLM Agents}

LLMs \cite{zhang2025scientific, schick2023toolformer} have advanced significantly, driving transformative impact across scientific tasks with strong processing and reasoning capabilities. This progress has spurred LLM-driven agents \cite{luo2025large}, revolutionizing complex scientific research automation via goal-directed behavior, dynamic decision-making, and environmental interaction. However, LLMs face inherent limitations in intricate computational tasks, especially those requiring precise numerical operations, symbolic manipulations, or multi-step deductions, due to reliance on fixed parametric knowledge \cite{mallen2022not, qu2025tool}, often leading to inaccuracies or ``hallucinations'' in high-stakes scientific computations \cite{zhuang2023toolqa}. Augmenting LLMs with external tools \cite{shen2024llm, lu2025octotools} thus emerges as a promising solution, enhancing stability and interpretability by grounding outputs in rigorous algorithms rather than probabilistic predictions.

Recent advances in integrating agents with external tools have boosted automated problem-solving \cite{gao2024confucius, wang2024what}. ReAct \cite{yao2023react} laid a foundation by unifying reasoning and acting but, focused on ``immediate acting'' and lacking long-term planning, limits handling complex multi-step tasks. Pre-Act \cite{rawat2025pre} refined this via pre-emptive planning: generating a global multi-step plan upfront reduces inefficiencies/errors from limited perspective and boosts efficiency, particularly for cross-step dependent tasks. OctoTools \cite{lu2025octotools} uses a hierarchical planning similar to Pre-Act and, as an extensible multi-agent tool framework, performs well across tasks. Yet current frameworks have limitations: (1) step-by-step validation may cause inefficiencies, redundancy, and overhead in stable tool collaboration; (2) tool dependency management focuses on explicit parameters but neglects critical implicit ones for multi-tool coordination. To address these, we propose Prog-Act and Resource-Pool to enhance multi-agent/multi-tool collaboration (details in Section \ref{sec:method}).


\begin{figure*}[t]
\centering
\includegraphics[width=0.9\textwidth]{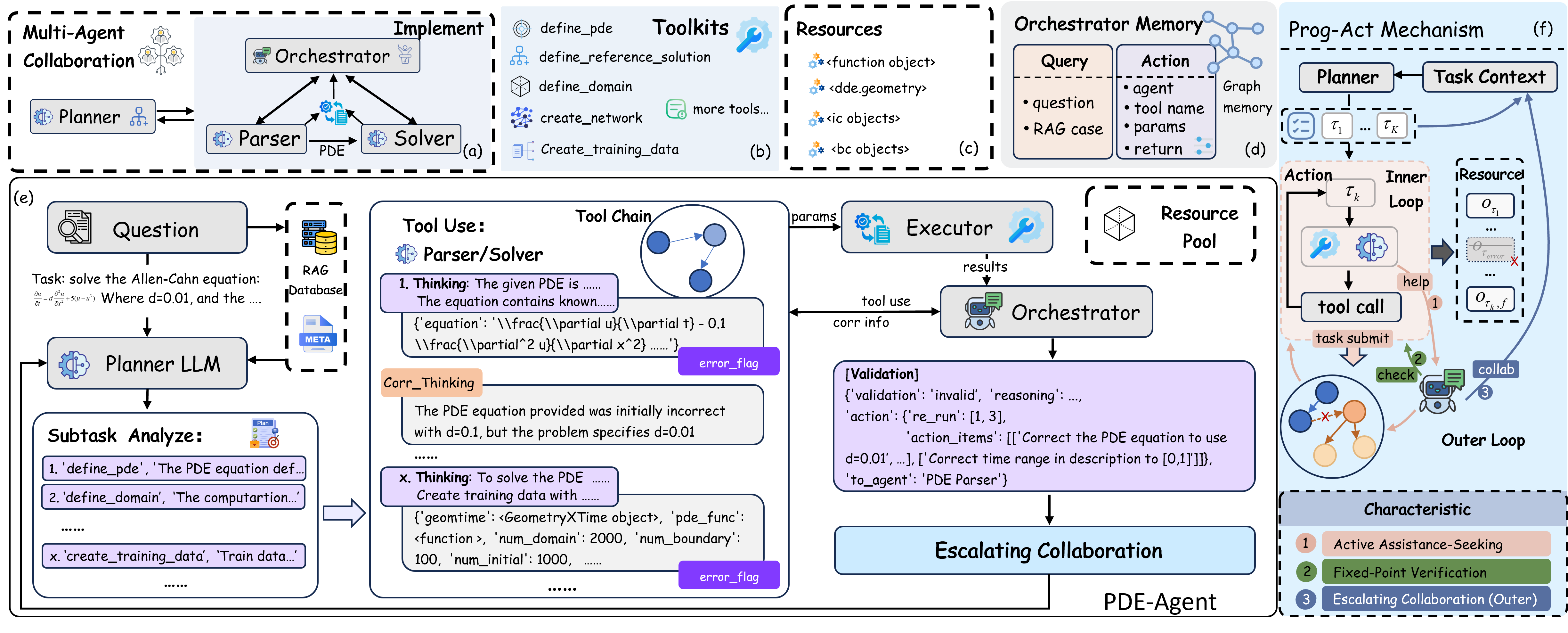} 
\caption{The architecture of PDE-Agent.}
\label{fig2}
\label{fig:PDE-Agent_architecture}
\end{figure*}

\section{Method: PDE-Agent}
\label{sec:method}
To enable end-to-end automated PDE solving, we propose PDE-Agent, a toolchain-augmented multi-agent framework for PDE solving. The framework aims to realize full automation of the PDE solving, encompassing problem understanding, task decomposition, tool collaboration, and dynamic validation, while integrating specifically designed mechanisms (Section \ref{sec:mult-agent_collaboration} and \ref{sec:multi-tool_coordination}) to optimize these types of complex, multi-step processes. Its effectiveness is validated using our PDE-Data and multi-level evaluation metrics.


\subsection{Overall Architecture of PDE-Agent}

The PDE-Agent framework comprises four core components, as illustrated in Fig.\ref{fig:PDE-Agent_architecture}: Planner, Parser/Solver, Executor and Orchestrator. These components are designed to operate in a closed-loop process, collectively enabling end-to-end PDE solving from natural language input. 

Generally, user queries $\mathcal{Q}$ for PDE problem are typically unstructured and often embody complex intentions. Thus, the Planner first conducts a comprehensive analysis of the PDE problem and solution requirements to generate a holistic plan, decomposing the problem into potential executable $K$ subtasks denoted as $\mathcal{T}=\{\tau\}_{k=1}^K$ . This decomposition ensures consistency in long-term task planning. For each subtask $\tau=\langle k, f,g \rangle \in \mathcal{T}$, The Planner specifies: a task ID $k$, a target tool $f\in\mathcal{F}$ (where $\mathcal{F}=\{f\}_{n=1}^{N}$ denotes the set of available $N$ tools), and a subtask-specific goal/reasoning $g$. 

Subsequently, the Parser/Solver and Executor iteratively handle each subtask, with results aggregated by the Orchestrator. To stabilize tool utilization, we adopt a tool-parameter separation mechanism: the Parser/Solver verifies tool validity, extracts task-specific parameters $\theta_{\tau,f}$ for subtask $\tau$ or tool $f$.  (with the Parser extracting critical PDE objects/conditions like domain or boundary/initial conditions, and the Solver configuring neural networks to perform core solving), while the Executor invokes tool $f$ using extracted parameters $\theta_{\tau,f}$, , executes it, and outputs results.

After each subtask, the action $a:=\{ k, f, \theta, o \}$ is synchronously submitted to the Orchestrator. As ``Cognitive Elite'' of PDE-Agent, the Orchestrator oversees the entire task lifecycle, with primary responsibilities including validating the accuracy of subtask executions and implementing error correction such as flagging invalid parameters or reconfiguring tools for failed subtasks. Upon completion of all subtasks, the final solution to the PDE problem is formally represented as: 
\begin{equation}
    \mathcal{S} = Orchestrator^{*}(\mathcal{Q},\{ Executor(\tau , f, \theta_{\tau, f}) \}_{\tau \in \mathcal{T}})
\end{equation}


\subsection{Multi-agent Collaboration}
\label{sec:mult-agent_collaboration}

\textbf{Is Step-by-step Validation Necessary}? In our multi-agent architecture of PDE-Agent, ensuring coherence across multi-step task and maintaining global control over PDE solving require a holistic planning approach the Planner generates. To enhance adaptability to external feedback, existing methods adopt step-by-step validation to enable dynamic replanning; admittedly, this augments the agent’s reasoning and task adaptability. However, is step-wise verification truly necessary? Is such an overly cautious design warranted?

\textbf{Prog-Act}. Tool-augmented, LLM-driven agents simultaneously inherit the reasoning capacity of LLMs and the deterministic controllability of external tools, thereby exhibiting higher stability than the code-generation approach. Against this backdrop, step-wise validation risks introducing excessive information redundancy, leading to unwarranted consumption of temporal and computational resources. To address this trade-off, we propose the Prog-Act framework: a dual-loop error detection and correction mechanism that fully exploits multi-agent collaboration to balance adaptability and efficiency. Unlike previous step-by-step validation, Prog-Act introduces active assistance-seeking, fixed-point verification and escalating collaboration to optimize error handling while preserving task continuity, shown in Fig.\ref{fig:PDE-Agent_architecture}f. 

The Prog-Act approach operates through two interconnected loops (dual-loop error handling): 

Inner loop: localized error correction. The Orchestrator, as the central coordinator, receives action submissions, with the latest system-wide information. Rather than step-wise validating, it verifies at fixed checkpoints (e.g., after critical tool invocations or predefined subtask counts); additionally, Parser/Solver and Executor can proactively request validation via active collaboration. This establishes systematized multi-agent division of labor, aligning with human cognitive intuition. During verification, the Orchestrator analyzes current action sequence  $\mathcal{A} = \{ a \}_{m=1}^{M}$ against the original PDE problem $\mathcal{Q}$ for targeted error correction.
\begin{equation}
    \mathrm{Validate}(\mathcal{Q}, \mathcal{A}_c,\mathcal{G})= 
	\begin{cases}
		\mathrm{Pass}&\mathrm{if} \operatorname{no Error}(\mathcal{A}_c),  \\
		\text{InLoop}(\mathcal{A}_c,\mathcal{G})&\text{otherwise}
	\end{cases}
\end{equation}
where $\mathcal{A}_c \subseteq\mathcal{A}$ is the action subset at checkpoint $c$, and $\mathcal{G}$ represents the graph memory detailed in next part. 

Outer loop: global plan revision. If inner-loop ``collapse'', the Orchestrator formally challenges the Planner, initiating global plan revision. The Planner delivers adaptive plans through continuous refinement or, when necessary, executes full replanning. This dual-loop approach ensures efficient handling of both transient errors (via the inner loop) and systemic issues (via the outer loop).
\begin{equation}
    \mathcal{T}_{t+1}=\mathrm{Planner}(\mathcal{Q},\mathcal{T}_t,\mathrm{Feedback}(\mathcal{A},\mathcal{G}))
\end{equation}
where $\mathrm{Feedback}(\mathcal{A},\mathcal{G})$ aggregates error and recovery actions from the inner loop. 


\textbf{Graph Memory}. To enhance error traceability and cross-step coordination, PDE-Agent employs a Graph Memory structure that models tool/subtask dependencies and data flows, further augmenting the Pre-Act as the icing on the cake. During execution, the system constructs a directed graph $\mathcal{G}=(\mathcal{V},\mathcal{E})$ , where nodes $v \in \mathcal{G}$ represent tools or subtasks, and directed edges $e \in \mathcal{E}$ represent data transmission or dependency relationships. The Orchestrator, with graph memory, can  immediately identify the tool or subtask nodes tainted by erroneous data and initiates subsequent remediation upon error localization. This graph-based memory enables efficient error localization and recovery, aligning with the ``progressive'' nature of Prog-Act's validation strategy.

This dual-loop, graph-augmented approach enables PDE-Agent to balance efficiency and robustness, addressing the limitations of both overly cautious step-by-step validation and purely end-to-end verification. By leveraging multi-agent collaboration and structured memory, Prog-Act ensures coherent execution of complex, multi-step, and tool/subtask-dependent process in PDE solving.

\subsection{Multi-tool Coordination}
\label{sec:multi-tool_coordination}


Effective coordination of specialized tools is critical for complex multi-step tasks like PDE solving. However, existing tool-augmented methods predominantly rely on explicit parameter passing, extracting downstream tool parameters from problem descriptions or prior outputs and encoding them directly into commands, failing to handle implicit dependencies, such as runtime artifacts (e.g., function objects, class instances) from preceding tools that cannot be serialized, a key limitation in multi-tool collaboration unmanaged by current frameworks.

To bridge this gap, we introduce the Resource-Pool ($\mathcal{R}$), which originally serves as a repository for ``creations'' generated by tools integrated with a tool-parameter
separation. Remarkably, it shows strong extensibility, supporting flexible integration of any resources required by the system. As such, this component can also serve as a dynamic repository for shared resources, accessible to all components within the framework, and allowing the Parser/Solver and the Executor to access and update interdependent resources. Formally, with the tool's results $o_{\tau, f}$ , the Resource-Pool is updated by: 
\begin{equation}
    \mathcal{R}\leftarrow\mathcal{R}\cup\{o_{\tau, f}\}
\end{equation}

By centralizing implicit resources management, the Resource-Pool enables seamless multi-tool coordination, aligning naturally with the tool-parameter separation architecture to address a critical limitation of existing frameworks.

\section{Benchmark and Evaluation Metrics}
In this section, we introduce our PDE dataset (PDE-Data) and details of the multi-level evaluation metrics we propose to assess multi-agent and multi-tool coordination effectiveness in automated problem solving.

\subsection{PDE-Data}

\textbf{Design Motivation}. Existing PDE datasets (e.g., PDEBench \cite{takamoto2022pdebench}) primarily serve the training and validation of numerical solving mothods, concentrating on the ``equation formulation - solution data'' mapping. However, they lack support for the full workflow of ``natural language problem description $\rightarrow$ tool-collaborative solving''. 
To address this gap and the evaluation requirements of PDE-Agent, we constructed PDE-Data, aiming to  provide a test dataset covering diverse PDE problems with full workflow annotations. It is manually curated, with detailed examples provided in the Appendix, enabling comprehensive evaluation of entire pipline: task completation $\rightarrow$ tool collaboration $\rightarrow$ tool parameters and returns.


\textbf{Data composition}. PDE-Data includes 100 differential equation test cases, with detailed distribution in Table \ref{tab:overall_performances}. 
Distributed to balance fundamental representativeness and scenario diversity, its ``foundational cases as mainstay, specialized scenarios as supplement'' composition supports assessing agents' core PDE-solving capabilities and generalization in complex/niche scenarios, providing comprehensive multi-dimensional evaluation data.

\subsection{Evaluation Metrics}
\label{sec:evaluation_metrics}

%
We propose three categories of evaluation metrics to comprehensively assess PDE-Agent: global task completion (a global metric measured via Pass@k or task success rate); local tool invocation assessment (a detailed evaluation of tool invocation specifics, including Semantic Textual Similarity (STS) and BERTScore for tool parameters and outputs); and logical collaboration process assessment. Detailed descriptions of the first two are provided in the Appendix. Here, we emphasize the third category: the newly proposed logical collaboration process assessment, a process-oriented evaluation that gauges the rationality of tool interaction logic.

\textbf{Logical Collaboration Process Assessment}. Evaluating logical rationality of tool invocations, we focus on coherence of tool interactions, formalizing collaboration as a directed graph (tool chain) $\mathcal{G}$ with nodes $\mathcal{V}$ (tools) and directed edges  $\mathcal{E}$ (inter-tool data flow). Let $\mathcal{G}_p(\mathcal{V}_p, \mathcal{E}_p)$ be the ground-truth pattern graph (from PDE-Data) and $\mathcal{G}_e(\mathcal{V}_e, \mathcal{E}_e)$ be PDE-Agent's actual execution graph. Assessment has two core components: (1) Tool chain completeness, scored via (partial) subgraph isomorphism and PageRank to preserve critical tools/data flows; (2) Tool chain error rate, quantified via graph edit distance and similarity to measure deviations from expected logic.


\textbf{(Partial) Subgraph Isomorphism-Based Evaluation}. This metric combines node-edge matching and partial subgraph ($\mathcal{G}_{sub}$) structural similarity to assess logical integrity: 

Node and edge matching: Evaluates the completeness of tools and data flows. It quantifies the proportion of nodes (tools) and edges (data dependencies) in $\mathcal{G}_e$ that match their counterparts in $\mathcal{G}_p$ 
\begin{equation}
    R_{\mathcal{V}} = \frac{|\mathcal{V}_e \cap \mathcal{V}_p |}{|\mathcal{V}_p|} , \,\,\, R_{\mathcal{E}} = \frac{|\mathcal{E}_e \cap \mathcal{E}_p |}{|\mathcal{E}_p|} 
\end{equation}

Structural Similarity of Partial Subgraphs: Assesses the logical coherence of internal toolchains through connectivity metrics, role node similarity, critical path similarity, and execution order similarity. Additional details on these metrics and other evaluation criteria are provided in the Appendix.

\textbf{PageRank-JS similarity metric}. 
The proposed metric quantifies topological congruence by comparing PageRank \cite{chung2014brief} distributions via Jensen-Shannon divergence. Firstly, we get the PageRank vector $\mathbf{p}_k$, which captures the relative importance of each tool node.
Then, we compute the Jensen-Shannon divergence between distributions. 
\begin{equation}
    \begin{aligned}
    D_{\mathrm{JS}}(\mathbf{p}_{\mathcal{V}_p}\parallel\mathbf{p}_{\mathcal{V}_e})=\frac{1}{2}\left[D_{\mathrm{KL}}(\mathbf{p}_{\mathcal{V}_p}^{\mathrm{norm}}\parallel M) + D_{\mathrm{KL}}(\mathbf{p}_{\mathcal{V}_e}^{\mathrm{norm}}\parallel M)\right]
    \end{aligned}
\end{equation}
where $M=\frac{1}{2}(\mathbf{p}_{\mathcal{V}_p}^{\mathrm{norm}}+\mathbf{p}_{\mathcal{V}_e}^{\mathrm{norm}})$, $\mathbf{p}_{k}^{\mathrm{norm}}=\frac{\mathbf{p}_k}{\|\mathbf{p}_k\|_1}$ is the L1-normalize vector and $D_{\mathrm{KL}}$ is the Kullback-Leibler divergence. 

Finally, we can obtain the similarity metric
\begin{equation}
    \mathrm{Sim}(\mathcal{G}_p,\mathcal{G}_e)=1-\sqrt{D_{\mathrm{JS}}(\mathbf{p}_{\mathcal{V}_p}\parallel\mathbf{p}_{\mathcal{V}_e})}
\end{equation}


\textbf{Graph Edit Distance-Based Similarity}. We propose a systematic similarity metric based on Normalized Graph Edit Distance (NGED) to quantify similarity between tool-chain. 
We calculated the maximum possible edit distance $Base(\mathcal{G}_e,\mathcal{G}_p)$ and can get the NGED
\begin{equation}
\begin{aligned}
    Base(\mathcal{G}_e,\mathcal{G}_p) = |\mathcal{V}_e| + |\mathcal{E}_e| + |\mathcal{V}_p| + |\mathcal{E}_p|  \\
    NGED(\mathcal{G}_e,\mathcal{G}_p) = \min\left(\frac{\mathrm{GED}(G_1,G_2)}{\mathrm{Base}(G_1,G_2)},1\right)
\end{aligned}
\end{equation}

Then we use the exponential mapping to transform NGED to the similarity score 
\begin{equation}
    S(\mathcal{G}_e,\mathcal{G}_p)=1-\frac{1-e^{-\alpha\cdot\mathrm{NGED}(\mathcal{G}_e,\mathcal{G}_p)}}{1-e^{-\alpha}}
\end{equation}
where $\alpha > 0$ is a sensitivity parameter controlling the steepness of the similarity curve and we use $\alpha=2$ in our experiments. 


\textbf{Graph embedding similarity based evaluation}. 
%
The Graph Embedding Similarity employs the Node2Vec algorithm \cite{grover2016node2vec} to generate latent representations of tool nodes, and compute a graph-level embedding $\mathbf{g}_{\mathcal{G}}$ through centrality-weighted aggregation of node embeddings. 

Then we compute the raw similarity using cosine similarity and transform the raw similarity into a normalized score $S \in [0, 1]$ using a parameterized sigmoid transformation
\begin{equation}
\begin{aligned}
    \mathrm{sim}_{\mathrm{raw}}(\mathcal{G}_e,\mathcal{G}_p)=\frac{\mathbf{g}_{\mathcal{G}_e}\cdot\mathbf{g}_{\mathcal{G}_p}}{\|\mathbf{g}_{\mathcal{G}_e}\|\cdot\|\mathbf{g}_{\mathcal{G}_p}\|} \\
S(G_A,G_B)=\frac{\sigma(k\cdot\mathrm{sim}_\mathrm{raw})-\sigma(-k)}{\sigma(k)-\sigma(-k)}
\end{aligned}
\end{equation}
where $\sigma(x) = \frac{1}{1 + e^{-x}}$ represents sigmoid function and $k$ is slope parameter controlling score distribution ($k = 1$ by default).




\section{Experiments}
\label{sec:experiments}

To validate the effectiveness of PDE-Agent in automatic PDE solving from natural language, we conduct comprehensive experiments on our PDE-Data. The experiments aim to : (1) Compare overall performance with methods from diverse perspectives; (2) Analyze performance through multi-dimensional metrics; (3) Verify the contributions of mechanism via ablation studies and case analyses. Note that detailed sample-wise statistics and analyses from the experiments are provided in the Appendix.

\subsection{Overall Performance Comparison}

\textbf{Experimental Setup}. We use PDE-Data, a curated dataset, balancing commonly studied PDE categories and less frequent types for robust evaluation of the framework's ability to handle diverse PDE scenarios. We compare PDE-Agent against three

\begin{itemize}
\item \textbf{OctoTools}: A recent open-source agent-toolbox framework designed to streamline multi-tool workflows in complex reasoning tasks, which has achieved strong performance across various complex tasks.

\item \textbf{OpenAlita}: An agent system characterized by minimal predefinition and maximal self-evolution, which attempts to construct functions via web-search (similar to code generation) for agent tool augmentation. 

\item \textbf{LLM-DS}: A pure LLM baseline (DeepSeek-v3) without explicit tool augmentation, relying solely on internal reasoning for PDE solving.

\end{itemize}
For all compared methods, we adopt the same LLM (DeepSeek-v3) and hyperparameters to ensure fairness. We adopt global evaluation metrics (primarily Pass@1) for overall performance, and conduct in-depth analyses of tool-augmented methods using our multi-level metrics.
Notably, we adapted OctoTools and OpenAlita with the Resource-Pool; without this adaptation, they would be unable to complete toolchain collaboration tasks.

\begin{table}[t]
\centering
\small
\setlength{\tabcolsep}{4pt} 
\begin{tabular}{@{}lrcccc@{}}
\toprule
\textbf{Category} & \textbf{Total} & PDE-A & Octo & O-Alita & LLM-DS \\ 
\cmidrule(lr){1-1} \cmidrule(lr){2-2} \cmidrule(lr){3-6}
 Heat Equation           & 18 & 17 & 14 & 14 & 11  \\
 Diffusion               &   6          &  4 &  5 &  2 & 2 \\
 Diffusion-Reaction      &    7         &  7 &  6 &  7 &  4 \\
 Wave Equation           & 19 & 17 &  7 &  4 &  6 \\
 Klein-Gordon            &   9          &  6 &  7 &  3 &  3 \\
 Advection               & 6 &  6 &  5 &  6 &  6 \\
 Burgers                 &  4           &  3 &  4 &  3 &  3 \\
 Laplace                 & 9 &  9 &  5 &  6 &  8 \\
 Poisson                 &  8           &  8 &  2 &  3 &  3 \\
 Allen-Cahn              & 6 &  6 &  4 &  6 &  5 \\
 Others                  & 8         & 7 & 3 & 5 & 5  \\
 \midrule %
Overall  & 100 & 90\% & 62\% & 59\% & 56\% \\
\bottomrule
\end{tabular}
\caption{Overall performance by PDE type and methods.}
\label{tab:overall_performances}
\end{table}

Table \ref{tab:overall_performances} presents the Pass@1 scores of PDE-Agent and baselines across PDE-Data. PDE-Agent achieves an overall success rate of 90\%, outperforming other methods by significant margins. There are some key observations: 
\begin{itemize}
\item The pure LLM baseline (LLM-DS) performs poorly due to its inability to leverage tools - critical for reliable symbolic abstraction (e.g., define\_pde) and complex computations, both of which are essential for PDE solving. 
\item OpenAlita's automatic tool construction functionality currently relies heavily on code-generation capabilities, with limited abstraction of tool functionalities. This hinders its performance beyond basic tool setup. 
\item OctoTools, which employs a general step-by-step validation framework without toolchain augmentation, struggles with two key limitations: (1) Failure to trace dependencies across long-range, cross-step tasks; (2) Information redundancy from step-wise validation, misinterpretation of task intent and premature task termination. 
\item PDE-Agent's superiority stems from two core mechanisms: the Prog-Act framework (enabling multi-agent collaboration to handle unexpected errors during solving) and the Resource-Pool (facilitating tool coordination through effective management of implicit parameters).  
\end{itemize}

\begin{figure*}[t]
\centering
\includegraphics[width=0.85\textwidth]{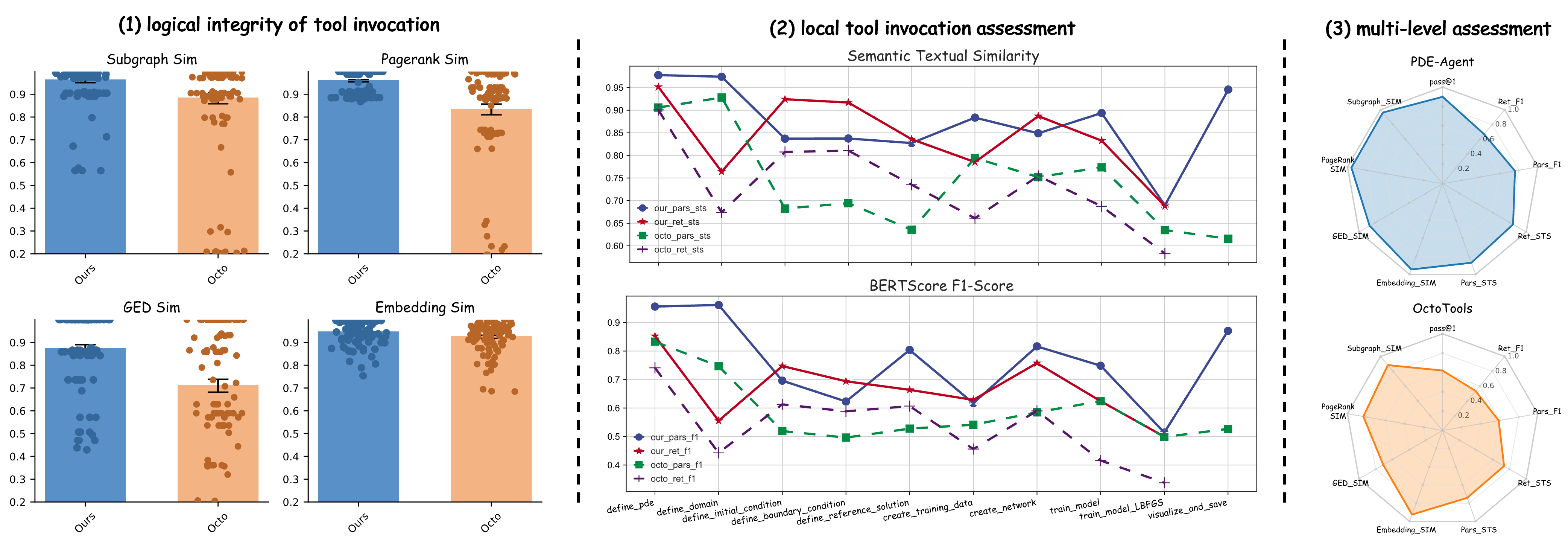} 
\caption{The multi-level evaluation.}
\label{fig:multi-level_assessment}
\end{figure*}

\subsection{In-Depth Comparison with OctoTools}

OctoTools is the most comparable baseline, as it focuses on tool-augmented, multi-step reasoning. However, overall success rates alone are insufficient to evaluate performance in multi-step tasks, as they fail to capture task completion quality, tool invocation logic, or step-by-step tool utilization. We thus conduct a fine-grained analysis across multiple distinct dimensions (Fig.\ref{fig:multi-level_assessment}). 

\textbf{Logical collaboration process assessment}. We evaluate the logical integrity of tool invocation performance using graph-based metrics which is detailed in Section \ref{sec:evaluation_metrics} which assess logical completeness and similarity.
The results of these metrics across all test cases (Fig.\ref{fig:multi-level_assessment}(1)) demonstrate that PDE-Agent outperforms OctoTools in all four dimensions. Notably, OctoTools exhibits moderate task completion quality despite lower overall success rates—an insight not captured by the global Pass@1 metric. More detailed results are provided in the Appendix.

\textbf{Local tool invocation assessment}. We measure Semantic Textual Similarity (STS) and BERTScore of the parameters and outputs for tool invocation to evaluate the accuracy of step-by-step tool utilization. STS can reflect alignment with task requirements. BERTScore, including precision, recall and F1-score, quantifies the semantic alignment.  As shown in Fig.\ref{fig:multi-level_assessment}(2), PDE-Agent achieves higher semantic similarity and BERTScore in both parameters and outputs, indicating more stable tool invocation and better mastery of tool functionalities. This confirms the effectiveness of its mechanisms in ensuring accurate, context-aware tool utilization. Furthermore, this assessment can serve two purposes: on one hand, it functions as an evaluation of the entire system; on the other hand, it enables researchers to conduct targeted optimization of tools or enhance agents' capabilities in thes specific aspect. Detailed evaluations and analyses are provided in Appendix.

The Fig.\ref{fig:multi-level_assessment}(3) summarizes multi-level evaluation metrics to reflect the overall performance of the methods. These results collectively demonstrate the PDE-Agent not only achieves higher overall success rates but also exhibits superior logical coherence in tool collaboration and more accurate local tool invocation.

\begin{table}[t]
\centering
\small
\setlength{\tabcolsep}{4pt} 
\begin{tabular}{@{}lcccccc@{}}
\toprule
\multirow{2}{*}{\textbf{Category}}  & \multicolumn{3}{c}{Prog-Act} & \multicolumn{3}{c}{diff-LLM} \\ 
\cmidrule(lr){2-4}  \cmidrule(lr){5-7}
 & Full & w/o & step & DS & o4-mini & Gemini \\
\cmidrule(lr){1-1}  \cmidrule(lr){2-4}  \cmidrule(lr){5-7}
 Heat Equation (18)             & 17 & 15 & 16 &   17 & 14 & 15 \\
 Diffusion (6)                  &  4 &  2 &  6 &    4 &  5 &  4    \\
 Diffusion-Reaction (7)         &  7 &  5 &  6 &    7 &  7 &  6     \\
 Wave Equation (19)             & 17 & 12 & 15 &   17 & 16 & 18  \\
 Klein-Gordon (9)               &  6 &  7 &  6 &    6 &  6 &  7  \\
 Advection (6)                  &  6 &  4 &  6 &    6 &  4 &  6     \\
 Burgers (4)                    &  3 &  3 &  2 &    3 &  3 &  3   \\
 Laplace (9)                    &  9 &  9 &  9 &    9 &  9 &  7   \\
 Poisson (8)                    &  8 &  5 &  6 &    8 &  4 &  5   \\
 Allen-Cahn (6)                 &  6 &  6 &  6 &    6 &  6 &  6   \\
 Others (8)                     &  7 &  8 &  8 &    7 &  7 &  8     \\
 \midrule %
Overall                         & 90\% & 76\% & 86\% & 90\% & 81\% & 85\%  \\
\bottomrule
\end{tabular}
\caption{Ablation study on the Prog-Act and different LLM for PDE-Agent.}
\label{tab:ablation_studies}
\end{table}

\subsection{Ablation Studies and Case Analysis}
The Resource-Pool design enables the propagation of runtime artifacts (e.g., function objects, class instances) between tools in collaborative workflows. In complex multi-tool tasks, effective information flow cannot be achieved without such centralized management. Thus, the Resource-Pool is indispensable for coherent tool collaboration.

To further clarify the role of the Prog-Act in PDE-Agent, we conducted ablation studies by disabling Prog-Act and comparing it with a step-wise dynamic validation strategy, contrasting it with the fully-equipped framework.

\textbf{Effectiveness of Prog-Act}. Table \ref{tab:ablation_studies} presents the performance of three PDE-Agent variants: the complete framework, PDE-Agent without Prog-Act (w/o Prog-Act), and a variant using step-wise validation (replacing Prog-Act’s dual-loop mechanism).

Removing Prog-Act reduces the success rate by 14\%. The framework reverts to a static tool chain, becoming unstable and unable to recover from errors (e.g., invalid parameter propagation). Compared to the step-wise validation variant, our framework achieves comparable performance while consuming significantly fewer resources. This confirms that Prog-Act’s dual-loop strategy (localized correction for transient errors and global revision for systemic issues) avoids the redundancy of step-wise validation, balancing efficiency and robustness.

\textbf{Generalizability Across Different LLMs}. We also tested its performance across different underlying LLMs to further validate the generalizability of PDE-Agent. Specifically, we evaluated PDE-Agent using three representative models: DeepSeek-v3 \cite{liu2024deepseek} (our primary test model), GPT-4o-mini \cite{hurst2024gpt} and Gemini \cite{comanici2025gemini}, maintaining consistent hyperparameters and tool configurations across all tests.
Table \ref{tab:ablation_studies} presents the overall success rates across these LLMs. 

\textbf{Case Analysis}. And we show the Prog-Act mechanism, encompassing active collaboration in the inner loop, global plan revision in the outer loop, alongside dynamic toolchain construction and error tracing.

\section{Conclusion}

In this work, we introduce PDE-Agent, a novel toolchain-augmented multi-agent collaboration framework for fully automated PDE solving, incorporating a Prog-Act framework with graph memory to enhance multi-agent collaboration and a Resource-Pool with tool-parameter separation to improve multi-tool coordination. We also developed PDE-Data, a multi-type benchmark with diverse PDE problems and annotations, and a multi-level metrics. These contributions advance automated PDE solving by providing a robust framework tailored to toolchain-augmented multi-agent challenges and standardized evaluation infrastructure, with experimental results validating PDE-Agent's superiority over existing methods in overall success and fine-grained tool utilization.

\clearpage

\bibliographystyle{unsrt}
\bibliography{reference.bib}

\newpage
\appendix

\end{document}